\documentclass[10pt,twocolumn,letterpaper]{article}

\usepackage[T1]{fontenc}
\usepackage[utf8]{inputenc}
\usepackage{lmodern}
\usepackage[margin=0.85in]{geometry}
\usepackage{amsmath,amssymb}
\usepackage{booktabs}
\usepackage{graphicx}
\usepackage{xcolor}
\usepackage[font=small,labelfont=bf]{caption}
\usepackage{tikz}
\usetikzlibrary{positioning,arrows.meta,fit,backgrounds}
\usepackage{pgfplots}
\pgfplotsset{compat=1.16}
\usepackage[colorlinks=true,allcolors=blue]{hyperref}

\usepackage{needspace}
\usepackage{enumitem}
\setlist{topsep=3pt,itemsep=3pt,parsep=0pt}
\usepackage{titlesec}
\titlespacing*{\section}{0pt}{1.7ex plus .2ex minus .1ex}{1.0ex}
\titlespacing*{\subsection}{0pt}{1.4ex plus .2ex minus .1ex}{0.7ex}
\titlespacing*{\paragraph}{0pt}{1.1ex plus .1ex}{1em}
\let\origsection\section
\renewcommand{\section}{\Needspace*{11\baselineskip}\origsection}
\let\origsubsection\subsection
\renewcommand{\subsection}{\Needspace*{8\baselineskip}\origsubsection}
\let\origparagraph\paragraph
\renewcommand{\paragraph}{\Needspace*{4\baselineskip}\origparagraph}

\definecolor{hidecol}{HTML}{2E7D32}
\definecolor{breakcol}{HTML}{C62828}

\newcommand{\gelu}{\mathrm{gelu}}
\newcommand{\softplus}{\mathrm{softplus}}
\newcommand{\Blocks}{\mathrm{Blocks}}
\newcommand{\KL}{\mathrm{KL}}
\newcommand{\R}{\mathbb{R}}

\title{\bf Privacy-preserving inference on hosted models:\\
a learned input-embedding transform for frozen Qwen3.8-27B}

\author{
\begin{tabular}{c@{\hskip 2.2cm}c}
Mahendra Roopa & Srikanth Roopa\\[2pt]
Spinal & Spinal\\[2pt]
\texttt{mahendra@getspinal.com} & \texttt{srikanth@getspinal.com}
\end{tabular}
}
\date{20.10.2026}

\begin{document}

\makeatletter
\twocolumn[
\begin{@twocolumnfalse}
\maketitle
\begin{center}\begin{minipage}{0.88\textwidth}
\begin{abstract}
Every prompt sent to a hosted language model is readable by whoever runs it.
We study whether the prompt can remain on the client while the host receives
transformed vectors that stand in for it. A language model converts text to
vectors exactly once, with a public lookup table; we perform that step locally
and pass the vectors through a small trained network (55.6M parameters) before
they leave the perimeter. The host runs a frozen 27B model (Qwen3.8-27B) on the transformed
vectors and never receives text or token ids.

We train the transform against the frozen model as a fixed judge, combining a
forward-KL utility term with an auxiliary term that pushes each vector away
from its original direction. On 1{,}490 MMLU questions held out from training, the transform
costs $1.5$ points of accuracy ($80.7\%$ vs.\ $82.3\%$), breaking $68$ of the
$1{,}226$ questions the unmodified model answers correctly ($5.5\%$), while
$30.1\%$ of content tokens no longer decode to their original token under a
nearest-neighbour lookup. We report sweeps over the inference-time strength
scalar and the training-time hiding weight, and two continuation budgets.
Neither sweep improves the trade: hiding collapses to $1.1\%$ at strength
$0.8$ while $48$ answers are still broken, and every hiding weight above $0.3$
roughly doubles the breaks. Continued training is the one exception, improving
both axes at $2{,}000$ additional examples on two seeds before reversing at
$6{,}000$.

Nearest-neighbour hiding is only a diagnostic; it is \emph{not} a privacy guarantee. An adversary that can run a transform of this family on inputs of its own choosing recovers the prompt outright. Estimating the displacement subspace at rank $64$ and projecting it out returns all $2{,}794$ content tokens of a held-out 96-prompt set from the checkpoint reported above.
The hiding rate on that same set reads $59.7\%$. Recovery is equally
complete in all 42 cross-checkpoint pairs tested, so keeping the transform's
weights on-prem does not help either. That an unkeyed transform of this family falls to an adaptive attacker is already known~\cite{osnip}. We additionally report the transfer between checkpoints and the utility cost measured alongside it. Accordingly, we do not claim privacy. The transform preserves much of the model's utility under the measurement protocol given here, and fails against a simple adaptive attack.
\end{abstract}
\end{minipage}\end{center}
\vspace{6mm}
\end{@twocolumnfalse}
]
\makeatother

\section{Introduction}

A hosted model needs the whole input to do its work. Whatever is sent, the host
holds a copy for at least the duration of the request. For sensitive inputs such as customer records, medical notes or internal documents, this means the provider sees the data during inference.

Each of the usual mitigations gives something up. Redaction changes the
question the model is asked, so it can change the answer that comes back.
Self-hosting removes the host from the picture, but the cost of serving a 27B
model then falls on the user. A trusted execution environment seals the input
inside the host's hardware~\cite{sgx}, which helps as far as the host's
configuration can be trusted.

This paper measures a different route. A transformer consumes text only through
its first step: tokenisation followed by a lookup in a public embedding table.
Every later layer is arithmetic on vectors. The lookup can therefore be moved to
the client, and the host given vectors instead of tokens. The table is public,
so the move alone hides nothing: anyone can invert a raw embedding by a table
lookup. We therefore measure whether a learned transform can displace those vectors far enough to frustrate recovery while the frozen model still computes the right answer.

Both properties have to hold at once, and either one alone is easy to get.
Neither extreme is useful: raw embeddings are trivially invertible, whereas noise strong enough to hide the input destroys the model's utility. The difficulty is in obtaining the two
together, which is what the rest of this paper measures. We do not obtain them:
the utility side is affordable and we characterise its price, and the privacy
side fails outright to an adversary that can run the transform on inputs of its
own choosing (\S\ref{sec:attack}).

\paragraph{Contributions.}
\begin{enumerate}\itemsep2pt
\item A contextual input-embedding transform (\S\ref{sec:method}) trained
against frozen Qwen3.8-27B, initialised exactly at the identity, with a
single inference-time strength scalar. It uses the same objective as~\cite{osnip}, applied here to Qwen3.8-27B (\S\ref{sec:related}).
\item A measurement protocol (\S\ref{sec:setup}) that separates utility from
obfuscation, pre-registers pass/fail conditions before each scored utility
run, and verifies
that an untrained transform changes nothing bit-for-bit.
\item Results on 1{,}490 MMLU questions held out from training (\S\ref{sec:results}),
including sweeps over inference strength, training hiding weight and
continuation budget, with per-question paired counts.
\item A chosen-input attack that recovers every content token of the
evaluated checkpoint at rank $64$ (\S\ref{sec:attack}). The recovery transfers across all 42 checkpoint pairs tested, so the victim's own transform is not required, and the hiding rate registers none of it.
\end{enumerate}

\section{Threat model and scope}
\label{sec:threat}

\paragraph{What is in scope.} The prompt, as it crosses the client boundary.
The host receives a floating-point tensor and generation settings. It gets no
text and no token ids, and it never runs the tokeniser. \S\ref{sec:attack} tests whether transmitting only transformed vectors is sufficient to conceal the prompt.

\paragraph{Out of scope.} The answer. The host generates it and can
read it. Any fact the answer discloses is disclosed to the host, so a request
whose output restates a sensitive field cannot hide that field from the host by
this mechanism. Prompt length is visible, as is the number of generated tokens
and the timing of the request.

\paragraph{Adversary.} We distinguish two. A \emph{passive} adversary observes
the transmitted tensor and knows the public embedding table. An \emph{active}
adversary additionally submits inputs of its own choosing and observes the
corresponding tensors, learning the transform's response on chosen plaintexts.
The hiding rate of \S\ref{sec:metrics} characterises the weakest passive
attack, nearest-neighbour decoding. What the active adversary achieves is
reported in \S\ref{sec:attack}.

\paragraph{Custody of the transform.} The transform runs on the client, so a
deployment may hold $\theta$ on-prem and never publish it. We treat that as an
operational control rather than part of the boundary, and assume the adversary
knows $\theta$. The attack of \S\ref{sec:attack} does not rely on knowing $\theta$. The reason is structural: the rows of $\Delta$ lie in the row space of $W_2$, at most $r = 1024$ of $d = 5120$ dimensions, and with $\varepsilon = 0$ the remaining dimensions carry $E$ unchanged. That subspace is what the
attack of \S\ref{sec:attack} estimates, and it is a property of the
architecture and its training recipe rather than of one checkpoint---which is
why the attack transfers between checkpoints (\S\ref{sec:attack}) and why an
adversary holding \emph{any} member of the family is as dangerous as one
holding $\theta$ itself.

We expect, but have not shown, that an adversary confined to observing ordinary
protected traffic can reach the same subspace without submitting anything: the
positions that still decode correctly---the majority, or utility would
fail---supply matched pairs $(Z_i, E_i)$ whose differences span it. The one
observer-side attack we implemented, on a 7B pilot, had to guess $E$ and
recovered nothing. Closing that gap would remove the last operational
restriction on the adversary, and we flag it as untested rather than
unavailable.

\paragraph{Not a cryptographic claim.} The transform has no key and admits no
$\varepsilon$. It is empirical obfuscation measured against specific decoders,
and nothing stronger~\cite{dwork}.

\section{Method}
\label{sec:method}

\subsection{Setup and notation}

Let $x = (x_1,\dots,x_n)$ be the prompt's tokens and
$E = \mathrm{table}[x] \in \R^{n \times d}$ the corresponding rows of the frozen
host model's input embedding table, with $d = 5120$ and $248{,}320$ rows. Let
$M(\cdot\,,y)$ denote the frozen host model's next-token distributions along a
reference continuation $y$, given input vectors in place of an embedded prompt.
The host model $M$ and its table are never modified. Only the transform is
trained.

\subsection{The transform}

The transform $T_\theta : \R^{n\times d} \to \R^{n\times d}$ maps the prompt's
embeddings through a low-dimensional contextual bottleneck of width $r = 1024$:
\begin{align}
h &= \Blocks(E W_1 + P), & h &\in \R^{n\times r},\\
\Delta &= \gelu(h W_3)\,W_2, & \Delta &\in \R^{n\times d},\\
\sigma &= \softplus\!\big(\gelu(h W_4)\,W_5\big), & \sigma &\in \R^{n\times d},\\
Z &= E + a\,\Delta + \varepsilon \odot \sigma, & \varepsilon &\sim \mathcal{N}(0, I),
\end{align}
with $W_1 \in \R^{d\times r}$, $W_2, W_5 \in \R^{r\times d}$,
$W_3, W_4 \in \R^{r\times r}$, $P$ a sinusoidal position encoding, and
$\Blocks$ a stack of three pre-norm transformer encoder layers of width $r$
with 16 heads and feed-forward width 4096. Biases are omitted from the
equations for readability. The parameters
$\theta = \{W_1,\dots,W_5,\Blocks\}$ total $55{,}627{,}776$.

Because the blocks attend across positions, $\Delta_i$ depends on the whole
prompt, not on token $x_i$ alone: two occurrences of the same token in different
contexts are displaced differently. The noise scale $\sigma$ is produced in
embedding space by a second projection $W_5$, so the sampled perturbation has a
diagonal covariance in $\R^d$.\footnote{An alternative samples in the
$r$-dimensional latent space and shares $W_2$ for the projection, giving
correlated noise of rank $r$ and $50{,}379{,}776$ parameters. The checkpoint evaluated here uses the embedding-space form above. Every
checkpoint records which form it used.}

\paragraph{Pass-through positions.} Rows of special and formatting tokens are
transmitted unchanged, $Z_i = E_i$, since the host's chat template makes them public anyway. Content rows are the
only ones transformed, and the only ones scored.

\paragraph{Identity initialisation.} $W_2$ is zero at initialisation, so
$\Delta = 0$ and $Z = E$ exactly, and $\sigma$ is initialised at $10^{-3}$. The
untrained transform is therefore the identity function, and
$Z = E$ is an exact global optimum of the utility term alone: identical inputs
give identical logits, zero divergence and zero gradient. We verify
$Z = E$ bitwise before every run. Training moves $W_2$ off zero.

\subsection{Objective}

With $y$ the reference answer, the transform minimises
\begin{equation}
\begin{aligned}
L(\theta) = {}& \KL\big(M(E,y)\,\|\,M(Z,y)\big)\\
& + \lambda_c\, \operatorname*{mean}_i \big|\cos(E_i, Z_i)\big|\\
& + \lambda_n\, \operatorname*{mean}_i \big(\lVert Z_i\rVert - \lVert E_i\rVert\big)^2\\
& + \lambda_\sigma\, L_\sigma,
\qquad L_\sigma = -\operatorname{mean} \log \sigma,
\end{aligned}
\end{equation}
with $\lambda_c = 0.3$, $\lambda_n = 0.1$ and $\lambda_\sigma = 0.001$. The first term is a forward KL between the frozen model's answer distributions
under clean and transformed inputs, averaged over answer positions, in the
manner of distillation~\cite{distill}. The reference answer is appended to
both runs and never reaches the transform. The second term, which we call the
\emph{hiding} term, pushes each transmitted vector off its original direction.
The third holds vector length. The fourth rewards larger noise ($-\log\sigma$ is minimised by growing $\sigma$). At $\lambda_\sigma = 0.001$ against a utility weight of $1.0$ its contribution is negligible in our experiments. The
$\varepsilon$ of Eq.~(4) is sampled during training and set to zero at scoring,
following the usual reparameterised sampling construction~\cite{vae}.

\paragraph{Numerical precision.} The KL is accumulated in 64-bit precision. In
32-bit, softmax and log-softmax over a $248$k vocabulary do not cancel for
bitwise-identical logits: the loss reads $1.4\times10^{-8}$ instead of zero and
carries a $9.0\times10^{-8}$ gradient. AdamW normalises by the gradient's own
magnitude, so that residue is amplified into a full-size step and walks the
transform off the exact optimum described above. In 64-bit the same quantity is
$3.6\times10^{-17}$.

\subsection{Inference-time strength}

A trained transform is deployed with a scalar $a \in [0,1]$,
\begin{equation}
Z(a) = (1-a)\,E + a\,T_\theta(E),
\end{equation}
so $a = 0$ transmits the original vectors and $a = 1$ the fully transformed
ones. Training always uses $a = 1$. We vary $a$ at inference only
(\S\ref{sec:strength}). At evaluation time we verify numerically that $a = 0$ reproduces $E$ bitwise and $a = 1$ reproduces $T_\theta(E)$ bitwise.

\begin{figure}[t]
\centering
\begin{tikzpicture}[
  font=\scriptsize,
  box/.style={draw,rounded corners=2pt,align=center,inner sep=3pt,minimum width=2.1cm},
  >={Stealth[length=4pt]}
]
\node[box] (text) {prompt text};
\node[box,below=3.5mm of text] (tok) {tokenise};
\node[box,below=3.5mm of tok] (tab) {lookup table\\$248{,}320 \times 5120$};
\node[box,below=3.5mm of tab,fill=blue!6] (tr) {transform $T_\theta$\\$55.6$M, trained};
\node[box,below=13mm of tr,fill=gray!10] (host) {frozen Qwen3.8-27B\\64 layers};
\node[box,below=3.5mm of host,fill=gray!10] (ans) {answer text\\(not protected)};

\draw[->] (text) -- (tok);
\draw[->] (tok) -- (tab);
\draw[->] (tab) -- (tr);
\draw[->,thick] (tr) -- node[fill=white,inner sep=1.5pt,align=center]
      {$Z \in \R^{n\times 5120}$\\bf16 tensor, no tokens} (host);
\draw[->] (host) -- (ans);

\begin{scope}[on background layer]
\node[draw,dashed,rounded corners,fit=(text)(tok)(tab)(tr),inner sep=5pt,
      label={[font=\scriptsize\bfseries]above:CLIENT (trusted)}] {};
\node[draw,dashed,rounded corners,fit=(host)(ans),inner sep=5pt,
      label={[font=\scriptsize\bfseries]below:HOST (untrusted)}] {};
\end{scope}
\end{tikzpicture}
\caption{Where each step runs. Text never crosses the boundary; only the
transformed tensor does. The generated answer returns in the clear and is
readable by the host.}
\label{fig:pipeline}
\end{figure}
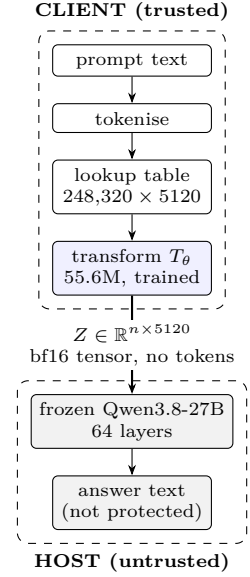

\section{Experimental setup}
\label{sec:setup}

\subsection{Model, data and optimisation}

The host model is Qwen3.8-27B, a frozen open-weight transformer
($d = 5120$, 64 layers, $248{,}320$ vocabulary rows) run in bfloat16 with a
pinned revision. The transform is trained on 800 instruction prompts from
Alpaca~\cite{alpaca}, all of at most 96 tokens, for 20{,}000 examples in
2{,}050 updates. Optimisation is AdamW~\cite{adamw} at
$\eta = 1.92\times10^{-5}$ with gradient clipping at $1.0$, on a single NVIDIA
H200. Training the transform requires a forward pass of the frozen model on both
clean and transformed inputs and a backward pass through the frozen model into
the transform; the frozen weights receive no update.

\subsection{Metrics}
\label{sec:metrics}

\paragraph{Hiding rate.} For each transformed content row we find the nearest
row of the public table and check whether it is still the original token:
\begin{equation}
\mathrm{hide}(i) = \mathbf{1}\big[\arg\min_j \lVert Z_i - \mathrm{table}[j]\rVert \neq x_i\big].
\end{equation}
We report the \emph{macro} rate, the mean over inputs of the per-input share, so
that each input counts equally, and also the \emph{pooled} rate over all content
tokens. This is a cheap test that can be run on every input and compared across
transforms. It tests one decoder. \S\ref{sec:limitations} sets out what it does not
establish.

\paragraph{Break rate.} Utility is scored on 1{,}490 MMLU
questions~\cite{mmlu}, four options each, spanning all 57 subjects, none of them
used in training. This is not an upstream MMLU split. It is carved from the
upstream \texttt{validation} split (1{,}531 items) by removing every item whose
group contains an item of the upstream \texttt{test} split, and by collapsing
duplicates within \texttt{validation} itself; 41 items are removed. Groups are
formed on two keys at once, the normalised question with its options sorted,
which catches permuted options, and the normalised question stem alone, which
catches a repeated question carrying different distractors. The published splits
overlap, and \texttt{test} is reserved here for a single final evaluation, so the exploration split gives up an overlapping group and the sealed split
keeps it. The carved file is frozen and hash-checked; the scorer refuses a
split whose hash does not match its manifest. Each question is asked twice, once from $E$ and once from $Z$, and
the model's chosen letter is recorded. Of the $n_0$ questions answered correctly
from $E$, the break rate is the share answered incorrectly from $Z$:
\begin{equation}
\mathrm{break} = \frac{|\{q : \mathrm{correct}_E(q) \wedge \neg\,\mathrm{correct}_Z(q)\}|}{n_0}.
\end{equation}
Questions that go from wrong to right are counted separately and are never
allowed to offset breaks. We therefore report repairs separately. We prefer this paired quantity to
token-level agreement with the clean generation, which has two defects: it rewards matching the clean wording even when both
answers are wrong, and it desynchronises after the first differing token.

\subsection{Protocol}

Each run's pass/fail conditions, budgets and analysis code are written and
frozen before the run, with their hashes recorded; post-hoc arms are labelled as
such in the artifact. Every scored run additionally asserts: the transform is
the identity at initialisation; the endpoints of Eq.~(6) are bitwise correct;
the host actually received the transformed tensor; the checkpoint hash matches
the launcher and the scored manifest; and a formatting gate on answer
extraction. Every number reported below was recomputed from the per-question rows by a separate reviewer model, and the differences that review produced are recorded with the run.
All of this describes the scored utility runs. The attack of
\S\ref{sec:attack} carried no pre-registered thresholds, and we report it as an
exploratory red-team result rather than a gated experiment.

\subsection{Artifacts}

Each run writes a directory holding its frozen conditions, the training log, the
per-question rows of every scored evaluation, the analysis script that produced
the reported quantities, and a manifest binding the scored checkpoint's SHA-256
to the launcher that trained it and to the transform configuration it was built
with. Every number in Tables~\ref{tab:main}--\ref{tab:continue},
Figure~\ref{fig:strength} and \S\ref{sec:attack} is read directly from those
files. Checkpoints are indexed by SHA-256 alongside the evaluation line of each run that scored them. Superseded and failed runs are kept in place.

At the time of writing these artifacts are internal. This paper neither
publishes them nor quotes the hashes that identify them, so a reader outside the
project can neither inspect the provenance described here nor reproduce
Tables~\ref{tab:main}--\ref{tab:continue}, Figure~\ref{fig:strength} or
\S\ref{sec:attack} from source. We record this as a limitation of the present
version.

\section{Results}
\label{sec:results}

\subsection{Accuracy cost}

Table~\ref{tab:main} gives the headline measurement. The transform costs $1.5$
accuracy points overall ($1{,}226$ correct answers become $1{,}203$). Per question, the change is larger than the aggregate suggests: $68$ correct answers become wrong, while $45$ wrong answers become correct.

\begin{table}[t]
\centering\small
\caption{Main result on 1{,}490 MMLU questions held out from training, $a=1$, noise off.}
\label{tab:main}
\begin{tabular}{lcc}
\toprule
& \textbf{Without} & \textbf{With} \\
& transform & transform \\
\midrule
Correct answers & 1{,}226 & 1{,}203 \\
Accuracy & 82.3\% & 80.7\% \\
Right $\to$ wrong (breaks) & --- & 68 \\
Break rate (of 1{,}226) & --- & 5.55\% \\
\quad 95\% interval & --- & 4.51--6.74\% \\
Wrong $\to$ right & --- & 45 \\
Tokens hidden, macro & 0\% & 30.08\% \\
Tokens hidden, pooled & 0\% & 24.85\% \\
\bottomrule
\end{tabular}
\end{table}

\subsection{Where the breaks fall}

Grouping the 1{,}226 clean-correct questions by the host model's decision margin
before transformation (the gap between the top option's score and the runner-up)
shows that breaks concentrate almost entirely in near-ties
(Table~\ref{tab:margin}). Confident answers are essentially untouched: 3 of 597.
The 45 repairs come from the same near-tie groups. Whether this holds outside
multiple-choice scoring is untested.

\begin{table}[t]
\centering\small
\caption{Breaks by the clean decision margin, in logit units.}
\label{tab:margin}
\begin{tabular}{lccc}
\toprule
\textbf{Margin} & \textbf{Questions} & \textbf{Broken} & \textbf{Share} \\
\midrule
$[0, 0.5)$ & 73 & 29 & 39.7\% \\
$[0.5, 1)$ & 85 & 17 & 20.0\% \\
$[1, 2)$ & 161 & 11 & 6.8\% \\
$[2, 4)$ & 310 & 8 & 2.6\% \\
$\geq 4$ & 597 & 3 & 0.5\% \\
\bottomrule
\end{tabular}
\end{table}

\subsection{Hiding depends on prompt length}
\label{sec:length}

The hiding rate is strongly length-dependent (Table~\ref{tab:length}): $41.5\%$
on prompts of at most 96 tokens against $13.0\%$ beyond 384. The transform was
trained only on prompts of at most 96 tokens, and on a held-out set of 96 short
Alpaca prompts of that kind---the \emph{anchor set}, used again in
\S\ref{sec:attack}---it hides $59.7\%$. Training length and hiding are associated here. We did not test whether the
association is causal, and the evaluation distribution differs from the
training distribution in more than length.

\begin{table}[t]
\centering\small
\caption{Macro hiding rate by prompt length, $a = 1$.}
\label{tab:length}
\begin{tabular}{lcc}
\toprule
\textbf{Prompt length} & \textbf{Inputs} & \textbf{Hidden} \\
\midrule
$\leq 96$ tokens & 564 & 41.5\% \\
97--192 & 666 & 26.1\% \\
193--384 & 216 & 16.1\% \\
$\geq 385$ & 44 & 13.0\% \\
\bottomrule
\end{tabular}
\end{table}

\subsection{Turning the strength down}
\label{sec:strength}

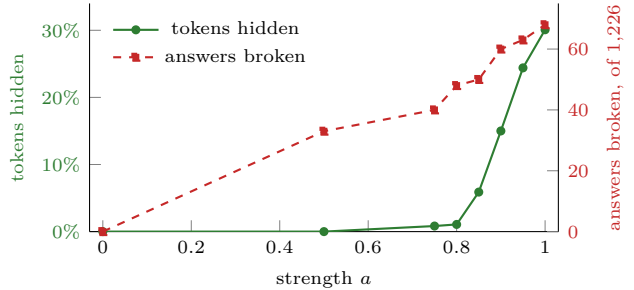
\begin{figure}[t]
\centering
\begin{tikzpicture}
\begin{axis}[
  width=0.92\columnwidth, height=4.6cm,
  xlabel={strength $a$}, xmin=-0.03, xmax=1.03,
  ylabel={tokens hidden}, ymin=0, ymax=34,
  yticklabel={\pgfmathprintnumber{\tick}\%},
  axis y line*=left, axis x line*=bottom,
  ylabel style={hidecol}, yticklabel style={hidecol},
  tick label style={font=\scriptsize}, label style={font=\scriptsize},
]
\addplot[hidecol,mark=*,mark size=1.3pt,thick] coordinates {
 (0,0) (0.5,0) (0.75,0.83) (0.8,1.07) (0.85,5.89) (0.9,15.0) (0.95,24.4) (1,30.08)};
\end{axis}
\begin{axis}[
  width=0.92\columnwidth, height=4.6cm,
  xmin=-0.03, xmax=1.03, ymin=0, ymax=75,
  axis y line*=right, axis x line=none,
  ylabel={answers broken, of 1{,}226},
  ylabel style={breakcol}, yticklabel style={breakcol},
  tick label style={font=\scriptsize}, label style={font=\scriptsize},
  legend style={at={(0.03,0.97)},anchor=north west,font=\scriptsize,draw=none,fill=none},
]
\addlegendimage{hidecol,mark=*,mark size=1.3pt,thick}\addlegendentry{tokens hidden}
\addplot[breakcol,mark=square*,mark size=1.3pt,thick,dashed] coordinates {
 (0,0) (0.5,33) (0.75,40) (0.8,48) (0.85,50) (0.9,60) (0.95,63) (1,68)};
\addlegendentry{answers broken}
\end{axis}
\end{tikzpicture}
\caption{The same transform and evaluation set at eight inference strengths.
Hiding (green, left axis) is near zero at $a \le 0.8$, by which point roughly two
thirds of the breaks (red, right axis) have already occurred.}
\label{fig:strength}
\end{figure}

The sweep shows no useful intermediate operating point (Figure~\ref{fig:strength}).
Hiding collapses long before the breaks do: at $a = 0.8$ only $1.1\%$ of tokens
are hidden while $48$ answers are still broken, and at $a = 0.5$ nothing at all
is hidden while $33$ answers are broken. The vectors have moved at these settings, but at $a = 0.5$ none has crossed into another token's cell.
Against a working reference of a break rate below $1\%$, no setting
other than $a = 0$ comes close.

\subsection{The hiding weight}

The hiding weight $\lambda_c$ is a training-time choice, so each value requires a
new transform. Table~\ref{tab:lambda} reports one run each at
$\lambda_c \in \{0.3, 0.5, 0.7, 1.0\}$ with every other setting held fixed,
including seed, batching and schedule. Every value above $0.3$ breaks roughly
twice as many answers ($1.97$--$2.29\times$) for 5 to 10 additional points of
hiding, and none reaches $50\%$ hiding on this evaluation. The four points are
not monotone in either axis: $\lambda_c = 1.0$ both hides more and breaks fewer
than $0.5$ and $0.7$. With one seed per point we cannot treat that ordering as
reliable. The sweep therefore supports only the weaker conclusion: no value of $\lambda_c$ tested improves hiding at a proportionate cost in breaks.

\begin{table}[t]
\centering\small
\caption{Training-time hiding weight, one run each. ``Price'' is additional
breaks per additional point of macro hiding, relative to $\lambda_c = 0.3$.}
\label{tab:lambda}
\begin{tabular}{lccc}
\toprule
$\lambda_c$ & \textbf{Breaks} & \textbf{Hidden (macro/pooled)} & \textbf{Price} \\
\midrule
0.3 & 68 & 30.08\% / 24.85\% & --- \\
0.5 & 144 & 36.56\% / 32.18\% & 11.7 \\
0.7 & 156 & 34.84\% / 30.32\% & 18.5 \\
1.0 & 134 & 39.59\% / 36.03\% & 6.9 \\
\bottomrule
\end{tabular}
\end{table}

\subsection{Continued training}

Continuing training on the same data with a restarted optimiser improves both
axes at a 2{,}000-example budget, on two seeds, but reverses at 6{,}000
(Table~\ref{tab:continue}): breaks rise to $102$ and $72$ against the parent's
$68$, at slightly higher hiding. The 6{,}000-example arms were a pre-registered test
that failed on both seeds. Because the optimiser was restarted, these arms do not isolate the effect of additional training examples. A single parent checkpoint underlies all four.

\begin{table}[t]
\centering\small
\caption{Continuation budgets from the same parent checkpoint.}
\label{tab:continue}
\begin{tabular}{lcc}
\toprule
\textbf{Arm} & \textbf{Breaks} & \textbf{Hidden (macro)} \\
\midrule
Parent & 68 & 30.08\% \\
$+$2{,}000 examples, seed A & 52 & 33.30\% \\
$+$2{,}000 examples, seed B & 55 & 30.61\% \\
$+$6{,}000 examples, seed A & 102 & 33.60\% \\
$+$6{,}000 examples, seed B & 72 & 34.31\% \\
\bottomrule
\end{tabular}
\end{table}

\subsection{The chosen-input attack}
\label{sec:attack}

The recovery reported here is conditional, and the conditions are worth stating
before the result. The adversary needs two things: a copy of a transform trained
by this recipe---not necessarily the victim's own---and the ability to run it on
inputs of its choosing. Both are ordinary in a client-side deployment, since the
transform ships to the client in order to be used, but they are conditions, and
an adversary with neither is outside what this subsection measures.

Given both, the method is direct. For the evaluated attack we use the same inference configuration as the main result: $a = 1$ and $\varepsilon = 0$. On inputs it chooses, the adversary knows $E$ exactly, so it observes $Z - E = \Delta$. It takes the leading
right singular vectors of the displacements it has collected, projects
intercepted traffic onto their orthogonal complement, and applies the
nearest-neighbour lookup of Eq.~(7). The displacement cancels and $E$ survives, because Eq.~(2) confines $\Delta$ to at most $r$ of $d$ dimensions and leaves the remaining $d - r$ directions of $E$ untouched.

Rank $64$ suffices. On the anchor set of \S\ref{sec:length}---96 held-out
Alpaca prompts---the attack returns $2{,}794$ of $2{,}794$ content tokens from
the checkpoint of Table~\ref{tab:main}, with no learned weights and no gradient
steps. The hiding rate of \S\ref{sec:metrics} reads $59.7\%$ on that same set,
measured on the same tensors.
The two numbers together show that nearest-neighbour hiding substantially overstates protection against this attack: the rate reports one decoder's failure, and a decoder that removes the displacement is not that decoder.

The adversary's copy need not be the victim's. Estimating the subspace from one
checkpoint and applying it to traffic protected by a different one recovers
every content token in all 42 (attacker, victim) pairs tested. The subspace
appears to be a property of the architecture and training recipe rather than of
a particular run, which is what removes on-prem custody of $\theta$
(\S\ref{sec:threat}) from the set of controls that would help.

The broader vulnerability of an unkeyed transform is already known. \cite{osnip}, whose objective is ours
(\S\ref{sec:related}), reports $100\%$ attack success against the unkeyed
ablation of its own method under an adaptive white-box attacker, and identifies
key-conditioning as the difference between that row and a defensible system.
The transform measured here is unkeyed, and this result is what that predicts.
In our evaluation a rank-64 subspace estimate is sufficient, and the attack transfers between checkpoints. Eq.~(2) explains the low-rank structure it exploits: the displacement lies in a subspace of rank at most $r = 1024$.

\paragraph{Scope of the claim.} One attack family was run. We did not implement
a learned inverter or iterative correction~\cite{vec2text} against this
checkpoint, and the passive adversary of \S\ref{sec:threat}---restricted to
observing ordinary traffic, with no inputs of its own---remains untested. The
result above therefore establishes that this construction fails against one
specific and cheap adversary, which is sufficient to withhold a privacy claim.
It is not a general statement that every attack succeeds. The utility measurements of \S\ref{sec:results} were not collected under a threat model this paper defends.

\section{Limitations and future work}
\label{sec:limitations}

\paragraph{The hiding rate is not a privacy measure.} It reports the failure of
one decoder: nearest-neighbour lookup in the public table, position by position.
A fixed rotation of the embedding space defeats that decoder completely while
preserving every pairwise relationship in the data, so a high hiding rate
licenses no conclusion. A low hiding rate does show exposure, but a high one
does not show protection. \S\ref{sec:attack} measures that gap directly:
$59.7\%$ hidden on the anchor set of \S\ref{sec:length}, whose every content
token the chosen-input adversary returns. We report the rate because it is cheap enough to run on every
input and comparable across transforms. We use it only as a decoder-specific
diagnostic.

\paragraph{The fixed displacement subspace is structurally vulnerable.} Under the deterministic inference setting evaluated here ($\varepsilon = 0$), \S\ref{sec:attack} is decisive against the residual form $Z = E + a\Delta$ at any setting of the weights: Eq.~(2) confines the displacement to at most $r$ dimensions, so $E$ passes through unchanged in at least $d - r = 4096$ directions and no amount of training moves it. Sampled noise is a separate case we have not tested, since $\varepsilon \odot \sigma$ has diagonal covariance in $\R^d$ (\S\ref{sec:method}) rather than lying in the row space of $W_2$. Avoiding this attack appears to require either a displacement spanning $\R^d$ or a secret the adversary does not hold. We have evaluated neither. Keying is the route the literature has
taken~\cite{osnip}, and it relocates the problem rather than removing it: the
same ablation that shows the keyed model resisting a mismatched-key attacker
shows it yielding completely to one holding the right key, so what is being
defended is key management. Raising $r$ towards $d$ is the cheaper of the two to
test here and is the first item of future work. Whether the utility cost
measured here survives either change is unknown. The inversion
literature (\S\ref{sec:related}) points the same way, though none of those
attacks is implemented here.

\paragraph{The answer is unprotected.} The host generates the reply and can read
it. A workload whose answer restates the sensitive field gains nothing for that
field.

\paragraph{Evaluation is multiple-choice.} The break rate scores a single letter
per question, which bounds the damage to a four-way decision and says nothing
about free-form generation quality. A pilot on 96 free-text prompts broke 3 of
70 correct answers, too few to conclude from; a larger blind-graded run is
outstanding.

\paragraph{Statistical scope.} One seed per point in
Table~\ref{tab:lambda}; two seeds per continuation budget in
Table~\ref{tab:continue}. Confidence intervals treat
questions as independent, which questions within a subject are not, so the
stated intervals are optimistic. The paired tests between arms are exploratory
and gate no decision.

\paragraph{Distribution.} The transform saw only Alpaca instruction prompts of
at most 96 tokens. Hiding degrades with length (Table~\ref{tab:length}), and the
evaluation distribution differs from training in domain as well as length.

\paragraph{Scale.} Every number here comes from one small configuration: a
$55.6$M-parameter transform with a bottleneck of $r = 1024$, trained on 800
Alpaca prompts for $20{,}000$ examples on a single H200. Three scaling questions
remain. A corpus matched to the evaluation
distribution would separate the length dependence above from the training set
that produced it. A larger compute budget would allow more seeds and
training schedules. A wider bottleneck bears on \S\ref{sec:threat}
directly, since the rank bound on the displacement is set by $r$: the question
is whether subspace estimation still recovers the input as $r$ approaches $d$,
which a higher hiding rate does not answer. The one scale result we have
settles none of the three, since the continuation arms improved both axes at
$2{,}000$ additional examples and reversed at $6{,}000$
(Table~\ref{tab:continue}) with the optimiser restart confounded with the extra
examples.

\section{Related work}
\label{sec:related}

\paragraph{Cryptographic and hardware routes.} Confidential computing seals the
input inside host hardware, which moves the trust to the hardware vendor and
leaves the data itself unchanged~\cite{sgx}. Secure multi-party and homomorphic
protocols give provable guarantees at a cost in latency that is still the
binding constraint for models of this size; \cite{survey} surveys the practical
trade-offs across this space. Our method provides no comparable cryptographic or hardware-backed guarantee, and is not an alternative to these routes.

\paragraph{Inversion attacks.} These are the adversarial baseline. Embeddings
leak substantial input content to learned inverters~\cite{songrag}, iterative
correction recovers most short texts exactly~\cite{vec2text}, and
vocabulary-matching against hidden states recovers text in the third-party
inference setting specifically~\cite{thomas}. Our nearest-neighbour hiding rate
(\S\ref{sec:metrics}) is weaker than all three. The attack of
\S\ref{sec:attack} assumes more than any of them---query access to a transform
of the family---but needs no training in exchange, and recovers everything.

\paragraph{Client-side perturbation of representations.} This is the family the
present transform belongs to: the client holds the embedding step and sends
perturbed vectors onward. \cite{textobfuscator} obfuscates word representations
while keeping them usable; \cite{snd} splits the model so the embedding layer
runs on the client, adds locally differentially private noise, and denoises the
output client-side; \cite{inferdpt} and \cite{dyntext} sanitise at the text
level for closed-box models.

These methods differ primarily in the guarantees they provide. \cite{snd} carries a local
differential privacy budget; \cite{obfuslm} derives its protection from
$(k,\epsilon)$-anonymity; \cite{covariant} states Rényi-DP guarantees for a
keyed permutation scheme. Those are claims of a different type from the one
available here, which is empirical obfuscation measured against named decoders
and nothing more. The transform in this paper belongs to the last group, and
\S\ref{sec:attack} is the cost of that position.

\paragraph{What these systems are evaluated against.} The evaluation norm in
this family is a passive decoder. \cite{snd} reports nearest-neighbour token
inversion and an attribute-inference attack; \cite{textobfuscator} reports
text-reconstruction attacks; \cite{obfuslm} reports nearest-search attacks
together with a trained inverter, which is a stronger decoder than any we run
against our own checkpoint; \cite{covariant} reports token-frequency matching
and a learned substitution-deciphering attack; \cite{osnip} reports
nearest-neighbour retrieval and vocabulary matching. An adaptive adversary
appears once in this set, in \cite{osnip}'s key ablation, and a chosen-input
subspace estimation appears nowhere. We do not read this as an oversight peculiar
to any one paper. \cite{survey} makes a related observation, that
methods of this class ``do not protect model parameters,
inference internals, or adaptive multi-query interactions''---which is the
adversary of \S\ref{sec:attack}.

\paragraph{Relation to OSNIP.} \cite{osnip} is the closest prior work. Its
objective is ours: a forward KL between a
frozen model's predictive distributions under clean and perturbed embeddings,
plus a penalty on $|\cos(h,z)|$, with the perturbation applied residually and
the norm held fixed---in their case by projecting back onto the sphere of radius
$\lVert h\rVert$, in ours by the penalty $\lambda_n$ of Eq.~(5), and in both
cases by backpropagating through the frozen host. Their hiding term is hinged at
a margin $\epsilon$ where ours is not, and they train on a mixture that is $80\%$
Alpaca and evaluate on MMLU among others. We claim no methodological novelty
over that construction.

The substantive difference is the key. OSNIP conditions the transform on a
user-specific secret and adds a diversity term that forces different keys to
produce geometrically separated outputs; the transform measured here is
unkeyed. \cite{osnip} directly evaluates the effect of key conditioning in its
ablation, which reports $100\%$ attack success against the no-key variant under
an adaptive white-box attacker, against $0\%$ for the keyed model under a
mismatched key.
\S\ref{sec:attack} is therefore an independent confirmation of that row rather
than a new finding, at 27B scale and with the attack named and costed---rank
$64$, one SVD, no learned weights---and with the structural reason it works
(Eq.~(2) bounds the displacement to $r$ of $d$ dimensions) made explicit.

Relative to \cite{osnip} we additionally evaluate cross-checkpoint transfer and report utility as paired answer changes. Our recovery needs no access to the victim's own transform, only to a sibling trained by the same recipe, which holds across all 42 checkpoint pairs tested and shows that on-prem custody of the victim's $\theta$ does not prevent this attack. \cite{osnip} reports
$98.49$--$101.48\%$ retained performance; we report the same kind of
configuration as $68$ broken answers against $45$ repaired ones on paired
questions (\S\ref{sec:results}), concentrated in near-ties
(Table~\ref{tab:margin}). Both readings can be true of one system, and
\S\ref{sec:metrics} explains why we think only the second is informative for a
user who cares which answers change. We also note that \cite{osnip}'s own
ablation records $100\%$ attack success against the keyed model when the
attacker holds the correct key. In that ablation, resistance depends on the
attacker not possessing the correct key.

\paragraph{Mechanics.} Our objective is a distillation-style divergence between
a frozen teacher's answer distributions under clean and perturbed
inputs~\cite{distill}, with reparameterised sampling~\cite{vae}.

\section{Conclusion}

A contextual transform on input embeddings can be trained against frozen
Qwen3.8-27B without touching the host model. On 1{,}490 MMLU questions held
out from training, the host scores $80.7\%$ through the transform against
$82.3\%$ clean, and $5.5\%$ of the answers it previously got right come back
wrong. At the same time $30.1\%$ of content tokens no longer decode to
themselves under a nearest-neighbour lookup. The damage concentrates in near-ties, and neither
the inference-time strength scalar nor the training-time hiding weight buys a
better trade: hiding disappears at $a \le 0.8$ while most of the answer damage
remains, and every larger hiding weight roughly doubles the breaks.

The utility cost is therefore modest and reasonably well characterised, and it
is not the binding constraint. An adversary that holds a transform of this
family---any member of it, not the victim's---and can run it on chosen inputs
recovers every content token of the evaluated checkpoint at rank $64$
(\S\ref{sec:attack}), in all 42 pairs tested. Against that adversary neither the
hiding rate nor on-prem custody of the weights is a control. The experiments therefore characterise the utility cost of this transform and give a protocol for measuring it. They do not support a privacy claim: the residual architecture does not protect the prompt from an adversary who can query it.

The mechanism is not obviously beyond repair, and the repair the field has
adopted is a secret key~\cite{osnip}. In its own ablation the keyed model
resists an attacker holding the wrong key completely and yields to one holding
the right key completely.
Resistance there depends on the attacker not possessing the correct key, not on
the geometry of the embedding space. A displacement spanning $\R^d$ rather than a fixed
$r$-dimensional subspace is the other direction, untested here. Whether either preserves the utility measured above is unknown. The next experiments should test a wider bottleneck, a corpus matched to the evaluation distribution, and a budget affording several seeds per point. The chosen-input attack of \S\ref{sec:attack}, with thresholds declared in advance, should be run against any checkpoint they produce before it is described as private.

\end{document}